\documentclass[10pt,twocolumn,letterpaper]{article}

\usepackage{cvpr}              

\usepackage{marvosym}

\definecolor{cvprblue}{rgb}{0.21,0.49,0.74}
\usepackage[pagebackref,breaklinks,colorlinks,allcolors=cvprblue]{hyperref}
\usepackage{float}

\def\paperID{7955} 
\def\confName{CVPR}
\def\confYear{2026}

\newcommand{\model}{DVGT\xspace}

\title{DVGT: Driving Visual Geometry Transformer}

\author{
  Sicheng Zuo$^{1, *}$ \quad
  Zixun Xie$^{1, 3, 4,*}$ \quad
  Wenzhao Zheng$^{1, *, \text{\Letter}}$ \quad 
  Shaoqing Xu$^{2, 3, \dagger}$ \\
  Fang Li$^{2, 3}$ \quad
  Shengyin Jiang$^{3}$ \quad
  Long Chen$^{3}$ \quad
  Zhi-Xin Yang$^{2}$ \quad
  Jiwen Lu$^{1}$
  \vspace{2mm} \\
  $^1$Tsinghua University \quad $^2$University of Macau \quad $^3$Xiaomi EV \quad $^4$Peking University \\
  Project Page: \url{https://wzzheng.net/DVGT}\\
  Large Driving Models: \url{https://github.com/wzzheng/LDM}
}

\renewcommand{\footnoterule}{%
  \kern -3pt
  \hrule width 0.9\linewidth height 0.4pt
  \kern 2.6pt
}

\usepackage{standalone}
\usepackage{pgfplots}
\usepgfplotslibrary{polar}
\pgfplotsset{compat=1.18}

\definecolor{mapanyorange}{RGB}{245, 133, 24}   
\definecolor{streamvggtpurple}{RGB}{148, 103, 189} 
\definecolor{cut3rgray}{RGB}{128, 128, 128}  
\definecolor{vggtblue}{RGB}{60, 100, 180}    
\definecolor{driv3rgreen}{RGB}{60, 160, 60}  
\definecolor{ourred}{RGB}{194, 89, 86}       

\begin{document}

\twocolumn[{%
\renewcommand\twocolumn[1][]{#1}%
\vspace{-12mm}
\maketitle
\vspace{-11.5mm}
\begin{center}
    \centering
    \includegraphics[width=0.98\linewidth]{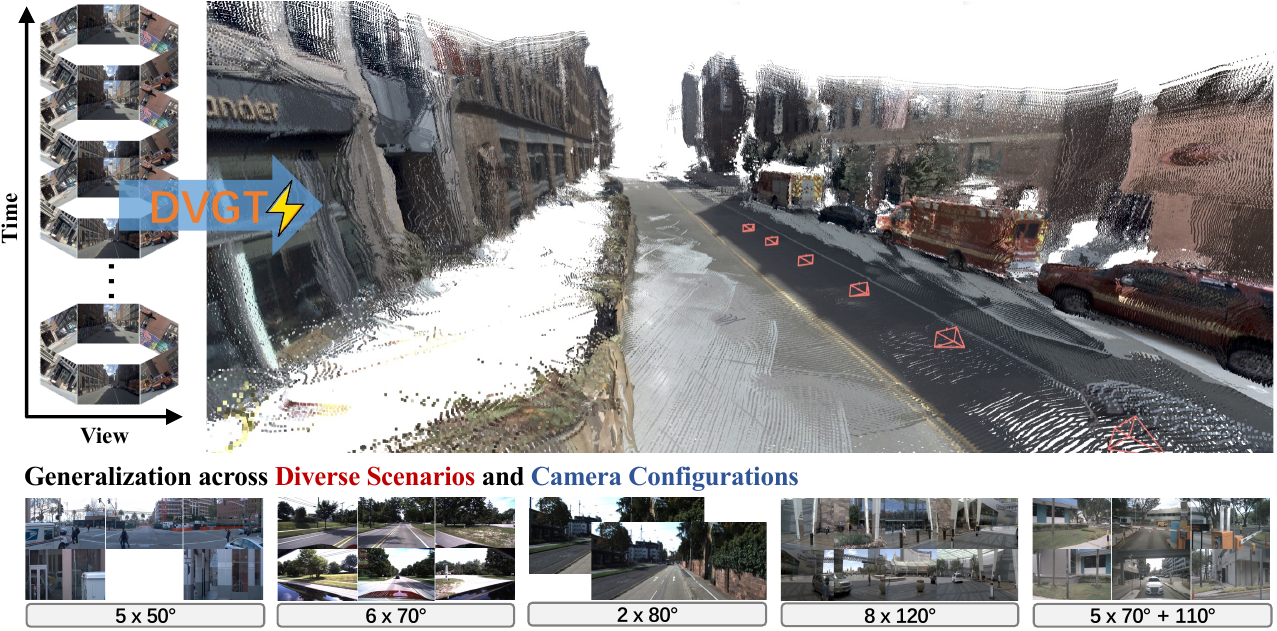}
    \vspace{-3mm}
    \captionof{figure}{
\textbf{\model is a large visual geometry transformer specifically designed for autonomous driving}. It accepts a sequence of unposed multi-view images and predicts a metric-scaled global 3D point map in the ego-centric coordinate system and the ego poses for each frame, which outperforms the other geometry prediction models without post-alignment with external sensors.
}
\label{teaser}
\end{center}%
}]

\begingroup
\renewcommand\thefootnote{}
\footnotetext{
  $^*$Equal contributions.
  $^\dagger$Project leader.
  \textsuperscript{\Letter}Corresponding author.
}
\endgroup

\begin{abstract}
Perceiving and reconstructing 3D scene geometry from visual inputs is crucial for autonomous driving. 
However, there still lacks a driving-targeted dense geometry perception model that can adapt to different scenarios and camera configurations.
To bridge this gap, we propose a Driving Visual Geometry Transformer (\model), which reconstructs a global dense 3D point map from a sequence of unposed multi-view visual inputs.
We first extract visual features for each image using a DINO backbone, and employ alternating intra-view local attention, cross-view spatial attention, and cross-frame temporal attention to infer geometric relations across images.
We then use multiple heads to decode a global point map in the ego coordinate of the first frame and the ego poses for each frame.
Unlike conventional methods that rely on precise camera parameters, \model is free of explicit 3D geometric priors, enabling flexible processing of arbitrary camera configurations. 
\model directly predicts metric-scaled geometry from image sequences, eliminating the need for post-alignment with external sensors. 
Trained on a large mixture of driving datasets including nuScenes, OpenScene, Waymo, KITTI, and DDAD, \model significantly outperforms existing models on various scenarios.
Code is available at \href{https://github.com/wzzheng/DVGT}{{\texttt{https://github.com/wzzheng/DVGT}}}.
\end{abstract}    

\section{Introduction}
\label{sec: intro}
Vision-centric autonomous driving has been widely explored due to its economic advantages and human-like sensing capabilities~\cite{bevformer, tpvformer, occworld, uniad, vad, doe-1, gaussianad, drivedreamer-1, vista, gaia-1, opendrivevla, orion, autovla}. 
Its cornerstone is the accurate perception of 3D scene geometry from visual inputs, where recent works have shown promising results with considerable potential~\cite{bevformer, tpvformer, occ3d, surroundocc, gaussianformer}.

Most existing methods focus on single-frame depth prediction~\cite{monodepth2, scenerf, visionnerf, mine, pixelnerf, surrounddepth} or 3D occupancy prediction~\cite{tpvformer, occ3d, surroundocc, occformer, gaussianformer, selfocc, occnerf, gaussianpretrain}, which require ground-truth poses for temporal fusion to achieve global 3D scene geometry understanding.
They usually rely on strong geometric priors and employ explicit 2D-to-3D projection to obtain 3D geometry~\cite{bevformer, bevdet, bevdepth, tpvformer, fb-occ, gaussianformer, opus}. 
This design choice tightly couples the model design to specific sensor configurations.
Also, most existing models are trained on limited data with a single camera configuration, hindering the adaptability and scalability across different vehicles and scenarios~\cite{dg-bev, unidrive}.
Although recent general visual geometry models~\cite{vggt,pi3,streamvggt,point3r} have shown strong 3D reconstruction performance, there still lacks a dense visual geometry model specifically designed for autonomous driving that can adapt to different scenarios and camera configurations.

\begin{figure}[t]
\centering
\vspace{1mm}
\resizebox{0.85\linewidth}{!}{
    \begin{tikzpicture}
    \begin{polaraxis}[
        width=10cm, height=10cm,
        clip=false,
        axis background/.style={fill=gray!3},
        ytick={0.2, 0.4, 0.6, 0.8, 1.0},
        yticklabels={0.2, 0.4, 0.6, 0.8, 1.0},
        grid=both,
        major grid style={gray!40, dashed, line width=1.0pt},
        minor grid style={gray!15, dotted},
        minor tick num=1,
        major tick style={draw=none},
        ymin=0, ymax=1.0, 
        xtick={90, 18, 306, 234, 162}, 
        xticklabels={
            \textbf{KITTI},
            \textbf{nuScenes},
            \textbf{Waymo}, 
            \textbf{OpenScene}, 
            \textbf{DDAD}
        },
        xticklabel style={
            rotate=\tick-90, 
            anchor=south
        },
        legend style={
            at={(0.5,-0.03)}, 
            anchor=north, 
            legend columns=3, 
            draw=none,
            /tikz/every even column/.append style={column sep=0.5cm},
            font=\footnotesize
        },
    ]


    \addplot+[color=mapanyorange, no markers, fill=mapanyorange, fill opacity=0.05, line width=1.2pt] 
    coordinates {(90,0.725) (18,0.269) (306,0.211) (234,0.240) (162,0.195) (90,0.725)};
    \addlegendentry{MapAnything}

    \addplot+[color=streamvggtpurple, no markers, fill=streamvggtpurple, fill opacity=0.05, line width=1.5pt] 
    coordinates {(90,0.469) (18,0.540) (306,0.584) (234,0.607) (162,0.415) (90,0.469)};
    \addlegendentry{StreamVGGT}

    \addplot+[color=cut3rgray, no markers, fill=cut3rgray, fill opacity=0.05, line width=1.5pt] 
    coordinates {(90,0.659) (18,0.547) (306,0.562) (234,0.593) (162,0.315) (90,0.659)};
    \addlegendentry{CUT3R}

    \addplot+[color=vggtblue, no markers, fill=vggtblue, fill opacity=0.1, line width=1.5pt] 
    coordinates {(90,0.801) (18,0.729) (306,0.811) (234,0.719) (162,0.476) (90,0.801)};
    \addlegendentry{VGGT}

    \addplot+[color=driv3rgreen, no markers, fill=driv3rgreen, fill opacity=0.1, line width=1.5pt] 
    coordinates {(90,0.784) (18,0.721) (306,0.770) (234,0.740) (162,0.740) (90,0.784)};
    \addlegendentry{Driv3R}

    \addplot+[solid, color=ourred, no markers, fill=ourred, fill opacity=0.15, line width=2pt] 
    coordinates {(90,0.849) (18,0.953) (306,0.921) (234,0.971) (162,0.837) (90,0.849)};
    \addlegendentry{\textbf{Ours}}

    \end{polaraxis}
\end{tikzpicture}
}
\vspace{-4mm}
\caption{\textbf{Quantitative comparison of 3D scene reconstruction.} Our method (red) demonstrates superior accuracy ($\delta < 1.25$ for ray depth estimation) across all evaluated datasets.
}
\label{fig:radarmap}
\vspace{-6mm}
\end{figure}
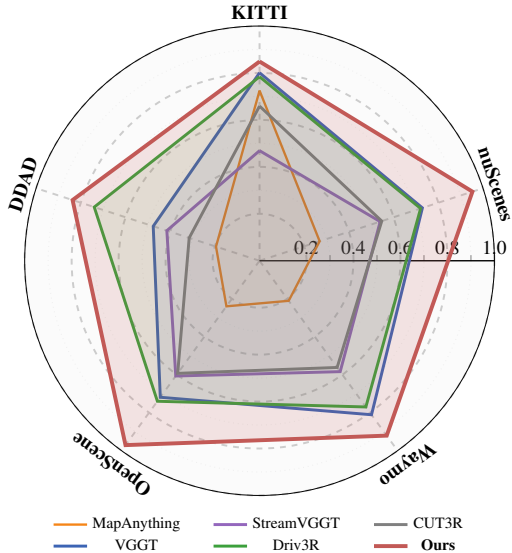

To bridge this gap, we propose a Driving Visual Geometry Transformer (\model) to generate global dense 3D point maps in an end-to-end manner, offering a continuous and high-fidelity geometric representation of the scene, as shown in Figure~\ref{teaser}. 
Given a sequence of unposed multi-view visual inputs, we first extract visual features for each image with a vision foundation model and then employ alternating intra-view local attention, cross-view spatial attention, and cross-frame temporal attention to infer geometric relations across images.
We add multiple heads to jointly predict the ego pose and the global 3D point map at the metric scale. 
Since the multi-view images are captured by fixed surrounding cameras in the ego car, we directly generate the global point map in the ego car coordinate of the first frame and predict the ego pose for each frame instead of the camera pose for each image.
Our model does not contain any spatial inductive bias and thus has more flexibility to adapt to different camera configurations.
For training, we construct a large-scale and diverse mixture of driving datasets, including nuScenes, OpenScene, Waymo, KITTI, and DDAD~\cite{kitti, nuscenes, waymo, openscene, ddad}, and generate dense geometric pseudo ground truths by aligning general-purpose monocular depth models~\cite{moge-2} with projected sparse LiDAR depth.
Extensive experiments show that our \model demonstrates superior performance over both general-purpose and driving-specific methods for 3D scene geometry reconstruction on various driving scenarios, as shown in Figure~\ref{fig:radarmap}. 
\model directly predicts metric-scaled global 3D point maps without the need for post-alignment with external sensors.




\section{Related Work}
\label{sec: related work}
\textbf{Geometry Perception for Autonomous Driving.} 
Early methods explored depth prediction from multi-view images~\cite{monodepth2, surrounddepth, scenerf, dist4d} to capture scene geometry.
Yet, these methods often provide a 2.5-D representation and struggle with producing a unified 3D scene.
To achieve a complete understanding of the scene geometry, TPVFormer~\cite{tpvformer} proposed 3D occupancy prediction, which uses dense voxels to describe the fine-grained geometry. 
Subsequent works have followed this direction, including voxel-based methods~\cite{openoccupancy, surroundocc, pointocc, occformer, occ3d} and object-centric methods~\cite{sparseocc, osp, opus, gaussianformer, gaussianformer2, gaussianworld, quadricformer}.
However, discrete occupancy grids introduce additional quantization errors (typically around 0.5m)~\cite{occ3d, openoccupancy}, which pose a challenge to represent geometry details accurately. 
In contrast, our \model predicts dense, continuous 3D point maps, enabling a fine-grained and complete representation of the scene geometry.

\textbf{Generalization Across Camera Configurations.}
The ability of driving perception models to generalize across different camera configurations is crucial for practical deployment. 
Conventional methods~\cite{bevformer, bevdet, tpvformer, gaussianformer} rely on explicit 2D-to-3D geometric projection to constrain the interaction between 2D images and 3D representations. 
This creates a strong dependency on camera-specific priors, which fundamentally limits their ability to generalize across different camera setups~\cite{dg-bev}. 
To mitigate this, recent work~\cite{unidrive} attempts cross-sensor generalization by projecting images into a unified virtual camera space. 
However, this approach remains dependent on geometric projection, and its performance degrades as camera parameters vary widely. 
In contrast, our proposed \model is a universal model that is adaptable to different camera configurations, which enables unified scaling on data from diverse sources.

\begin{figure*}[t]
\centering
\includegraphics[width=1.0\textwidth]{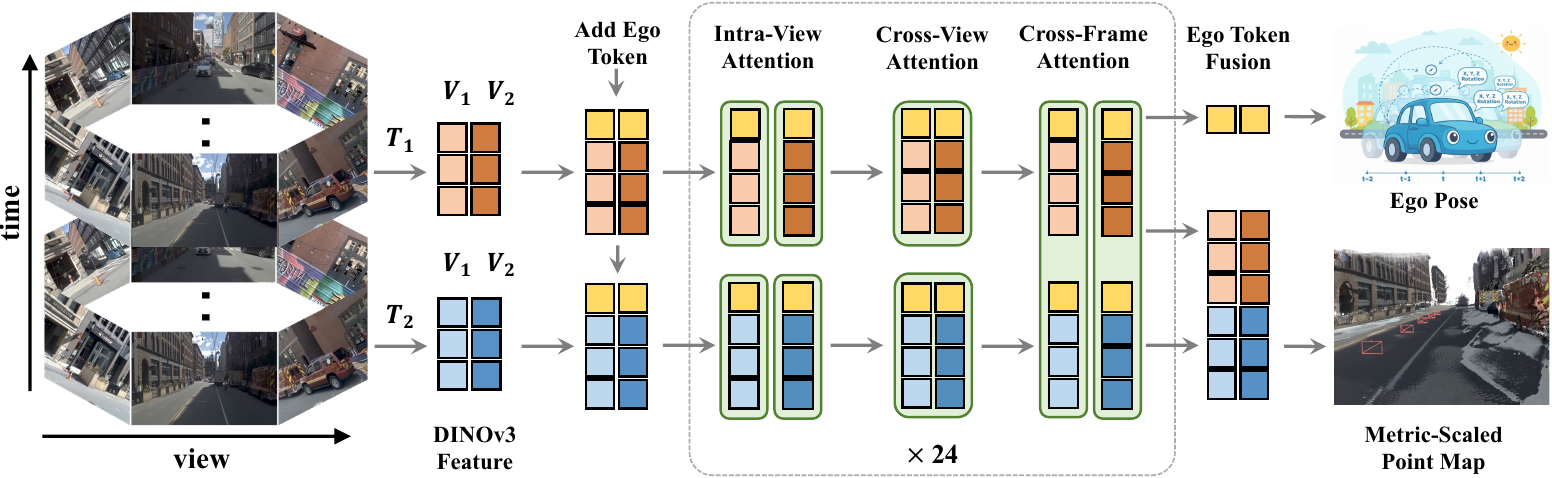}
\vspace{-7mm}
\caption{\textbf{Framework of our \model for metric-scaled 3D scene geometry prediction.} Given multi-frame, multi-view images, \model predicts metric-scaled 3D point map in the ego coordinate system of the first frame and the ego pose from each frame to the first frame.
}
\label{fig:framework}
\vspace{-5mm}
\end{figure*}

\textbf{General Visual Geometry Models.}
Recent general visual geometry models have shown significant progress, which reconstruct 3D point maps from images~\cite{dust3r, mast3r, spann3r, cut3r, streamvggt, point3r, vggt, pi3, mapanything}.
This paradigm, pioneered by DUSt3R~\cite{dust3r} for image pairs and advanced by following-up works~\cite{mast3r, cut3r, spann3r}, has been extended to multi-view reconstruction across various scenes by recent models~\cite{vggt, pi3}.
However, most models can only recover scene geometry at a relative scale and require post-alignment with external data like LiDAR points to obtain the metric scale of the scene~\cite{dust3r, spann3r, vggt, pi3, streamvggt}.
In contrast, our \model directly predicts metric-scaled scene geometry, which can be directly exploited in downstream tasks.
Also, existing models treat the multi-frame and multi-view images equally and estimate the geometry and pose for each image.
Instead, we employ a spatial-temporal architecture to efficiently process the driving inputs and directly generate the ego-aligned global point maps and ego poses in an end-to-end manner.

\section{Proposed Approach}
\label{sec: method}

\subsection{Ego-centric 3D Point Map Reconstruction}
\label{sec: method_task}

We first formulate the driving scene geometry perception as \textbf{ego-centric} 3D Point Map Reconstruction. 
This task involves predicting a continuous, metric-scaled 3D coordinate $\hat{\mathbf{P}}_{u,v} = (x, y, z)$ in a unified coordinate system for each pixel $(u, v)$ of an input image $\mathbf{I}$. 
The resulting dense and continuous 3D point map $\hat{\mathbf{P}} = \{\hat{\mathbf{P}}_{u,v}\}_{u=1..H,v=1..W}$, where $H, W$ are the image resolution, yields two key advantages: 
1. \textbf{High-Fidelity}: Its continuous coordinates eliminate quantization errors, enabling the precise modeling of scene geometry.
2. \textbf{Completeness}: Its pixel-aligned density ensures full coverage of visible regions, unifying the foreground objects and the background environment.

Conventional visual geometry models often reconstruct 3D point maps in a reference camera coordinate system~\cite{vggt, mapanything}. 
They tightly couple the model output to the reference camera intrinsics and extrinsics, hindering its adaptability to diverse sensor setups. 
To resolve this, we decouple the scene geometry representation from camera parameters by predicting 3D point maps in the ego-vehicle coordinate system of a reference frame. 
This ego-centric formulation yields a unified scene geometry representation that is invariant to camera focal lengths, poses, and the number of views, which is critical to a universal driving perception model. 

Formally, given an input image sequence $\mathcal{I} = \{\mathbf{I}_{t,n}\}_{t=1..T, n=1..N}$ from $T$ frames and $N$ views per frame, our model $\mathcal{M}$ jointly predicts the 3D point maps $\mathcal{P} = \{\hat{\mathbf{P}}_{t,n}\}_{t=1..T, n=1..N}$ and the ego-vehicle pose sequence $\mathcal{T}_{\text{ego}} = \{\hat{\mathbf{T}}_t\}_{t=1..T}$. 
Here, $\hat{\mathbf{P}}_{t,n} \in \mathbb{R}^{H \times W \times 3}$ is the point map for image $\mathbf{I}_{t,n}$, with all points expressed in the unified ego-vehicle coordinate system of the reference frame.
$\hat{\mathbf{T}}_t \in \text{SE}(3)$ is the ego motion from the reference frame to the frame $t$. 
The overall mapping is expressed as:
\vspace{-2mm}
\begin{equation}
    (\mathcal{P}, \mathcal{T}_{\text{ego}}) = \mathcal{M}(\mathcal{I}).
    \vspace{-1mm}
\end{equation}

\subsection{Spatial-Temporal Geometry Transformer}
\label{sec: method_model}
To achieve this goal, we propose \textbf{\model}, a universal driving visual geometry transformer. 
As shown in Figure~\ref{fig:framework}, our model consists of three components: an image encoder $\mathcal{E}$, a geometry transformer $\mathcal{F}$, and a set of prediction heads $\mathcal{H}$.

\textbf{Overall Framework}. 
Given multi-frame, multi-view image inputs, we adopt a pretrained vision foundation model~\cite{dinov3} to convert each input image $\mathbf{I}_{t,n}$ into a set of image tokens:
\vspace{-1mm}
\begin{equation}
    \mathbf{F}_{t,n} = \mathcal{E}(\mathbf{I}_{t,n}). 
\end{equation}
To facilitate pose prediction, we augment each image's tokens with a dedicated, learnable ego token $\mathbf{E}_{t,n}$:
\vspace{-2mm}
\begin{equation}
    \mathbf{Z}_{t,n} = \text{concat}([\mathbf{F}_{t,n}, \mathbf{E}_{t,n}]),
\vspace{-2mm}
\end{equation}
where $\text{concat}$ denotes the concatenation operation.
To distinguish tokens from different frames, all tokens are then augmented with temporal positional embeddings:
\vspace{-2mm}
\begin{equation}
    \mathbf{Z}'_{t,n} = \mathbf{Z}_{t,n} + \mathbf{Pos}_{t}.
\vspace{-2mm}
\end{equation}
where $\mathbf{Pos}_{t}$ denotes the temporal positional embeddings of the frame $t$.
The resulting sequence  $\mathcal{Z} = \{\mathbf{Z}'_{t,n}\}_{t=1..T, n=1..N}$ is then fed into the geometry transformer $\mathcal{F}$ and subsequently prediction heads $\mathcal{H}$ for joint 3D point map and ego pose prediction:
\vspace{-2mm}
\begin{equation}
    (\mathcal{P}, \mathcal{T}_{ego}) = \mathcal{H}(\mathcal{F}(\mathcal{Z})).
\vspace{-2mm}
\end{equation}

\begin{figure}[t]
\centering
\includegraphics[width=0.475\textwidth]{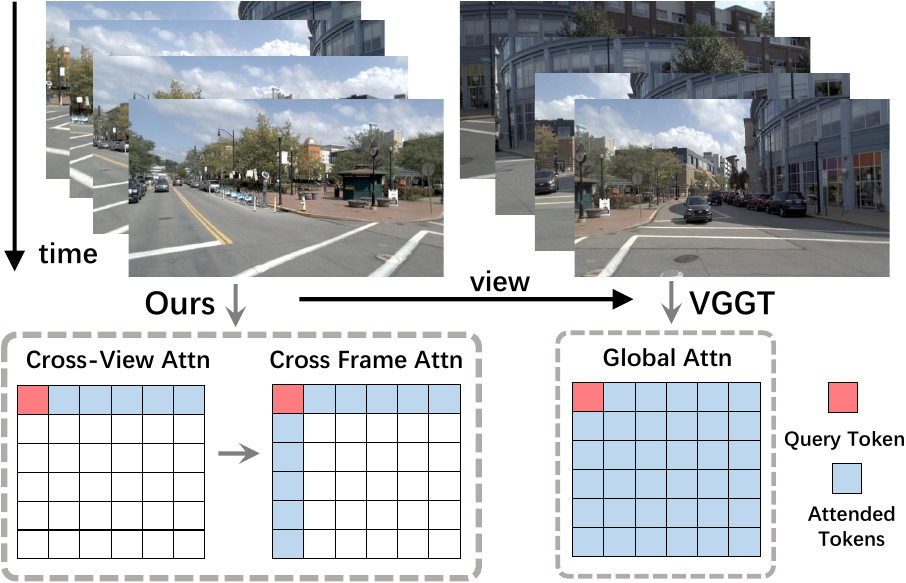}
\vspace{-7mm}
\caption{\textbf{Comparison of our factorized spatial-temporal attention (left) against the global attention used in VGGT (right).} 
}
\label{fig:method_attn}
\vspace{-5mm}
\end{figure}

\textbf{Geometry Transformer}. 
Most existing visual geometry models rely on global attention for interaction among all image tokens. 
They incurs substantial computational costs and are infeasible for real-time autonomous driving. 
To leverage the strong spatial-temporal structure in visual inputs of driving scenarios, we introduce an efficient factorized attention mechanism. 
Specifically, the geometry transformer $\mathcal{F}$ is composed of $L$ cascaded blocks, where each block sequentially performs three targeted attention operations:
\begin{itemize}
    \item \textbf{Intra-View Local Attention} operates exclusively on tokens within each image to refine local features.
    \item \textbf{Cross-View Spatial Attention} aggregates spatial information by attending to tokens across different views within the same frame.
    \item \textbf{Cross-Frame Temporal Attention} captures consistent statics and temporal dynamics by attending to tokens from the same view across different frames.
\end{itemize}
As shown in Figure~\ref{fig:method_attn}, this factorization efficiently factorizing the computationally expensive global attention while maintaining effective spatial-temporal information fusion.

\textbf{Prediction Heads}. 
The geometry transformer $\mathcal{F}$ outputs refined image tokens ${\mathbf{F}'}_{t,n}$ and ego tokens ${\mathbf{E}'}_{t,n}$. 
We then employ a 3D point map head $\mathcal{H}_{\text{point}}$ to decode the image tokens into the metric-scaled 3D point map: 
\vspace{-2mm}
\begin{equation}
    \hat{\mathbf{P}}_{t,n} = \mathcal{H}_{\text{point}}({\mathbf{F}'}_{t,n}). 
\vspace{-2mm}
\end{equation}
For pose prediction, the ego tokens from all views within the same frame are first aggregated into a global representation:
\vspace{-2mm}
\begin{equation}
    \bar{\mathbf{E}}_{t} = \sum_{n=1}^{N} {\mathbf{E}'}_{t,n}. 
\vspace{-2mm}
\end{equation}
This representation is then fed into a pose head $\mathcal{H}_{\text{pose}}$ to regress the ego pose of the frame $t$:
\vspace{-2mm}
\begin{equation}
    \hat{\mathbf{T}}_t = \mathcal{H}_{\text{pose}}(\bar{\mathbf{E}}_{t}).
\vspace{-2mm}
\end{equation}

\textbf{Prior-Free Design}. 
A core principle of \model is the 3D prior-free design. 
Unlike traditional driving perception models that rely on explicit camera parameters, our model is structurally independent of camera parameters and 2D-to-3D geometric projection designs. 
Instead, it learns to infer the 3D scene geometry directly from 2D image features in a purely data-driven, end-to-end manner. 
This fundamental design choice decouples the model from specific camera configurations, thereby granting it robust adaptability to diverse camera setups and driving scenarios.

\begin{figure*}[t]
\centering
\includegraphics[width=0.98\textwidth]{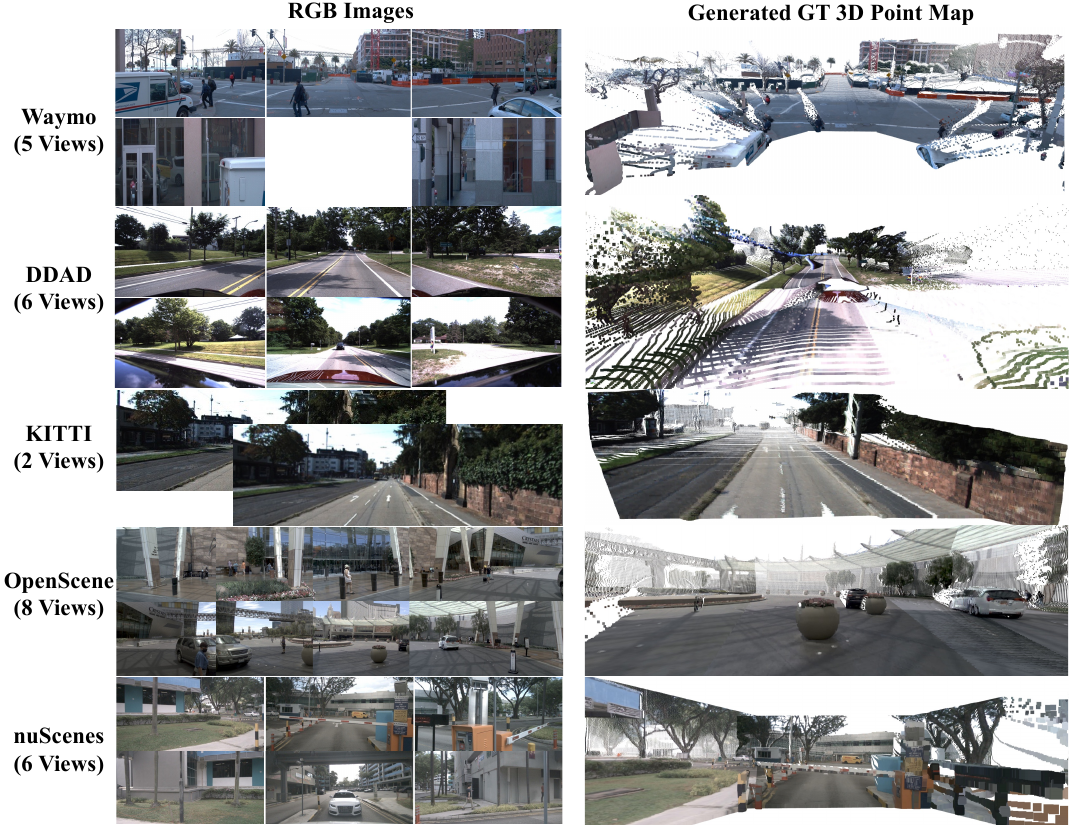}
\vspace{-3mm}
\caption{\textbf{We construct dense and accurate ground-truth 3D point maps for diverse driving scenarios.} This figure showcases examples from the Waymo~\cite{waymo}, nuScenes~\cite{nuscenes}, OpenScene~\cite{openscene}, DDAD~\cite{ddad}, and KITTI~\cite{kitti} datasets. For each scene, we display the multi-view RGB images (left) and the corresponding high-quality 3D point map (right), highlighting the diversity and precision of our data.
}
\label{fig:gt}
\vspace{-5mm}
\end{figure*}





\subsection{Dense Geometry Ground Truth Construction}\label{sec: method_data}
Accurate 3D reconstruction in autonomous driving is hindered by the scarcity of dense geometry ground truths. While aligning monocular depth predictions~\cite{metric3d, depthanything} with projected sparse LiDAR depth~\cite{omniscene} can generate pseudo-labels, it is often unreliable due to two primary factors: (1) the suboptimal generalization of general-purpose depth models in complex driving scenarios, and (2) the spatial unevenness of sparse LiDAR points, which often concentrates in small regions. These issues can lead to alignment failures or significant bias in estimated scale and shift parameters, resulting in grossly inaccurate pseudo-labels. To address this, we develop a robust pipeline to filter out samples where the depth model or alignment fails.

\textbf{Failure Pattern Analysis.} 
We conducted a rigorous analysis of failure cases in the pseudo-labeling process. We identified five dominant failure patterns: (a) \textbf{Semantic Misinterpretation:} Large, low-texture surfaces (e.g., truck trailers) are occasionally misclassified as sky. (b) \textbf{Photometric Instability:} Exposure issues can lead to stochastic depth predictions. (c) \textbf{Structural Ambiguity:} Complex planar patterns (e.g., billboards) are misinterpreted as having depth variance. (d) \textbf{Motion Artifacts:} Blur from high-speed driving or camera jitter degrades accuracy. (e) \textbf{Alignment Ill-Conditioning:} Extremely sparse or concentrated LiDAR points cause the alignment optimization to be ill-posed, resulting in significant errors for alignment.

\textbf{Threshold Filtering.} 
We evaluated multiple methods and selected MoGe-2~\cite{moge-2} for depth annotation and ROE~\cite{moge-1} for alignment. We established thresholds based on metric statistics to filter out low-quality pseudo-labels. Our strategy includes: (1) Valid Point Overlap: We compute the ratio of valid LiDAR points that are also predicted as valid by the depth model. This effectively identifies and removes samples suffering from patterns (a) and (b). (2) Standard Depth Metrics: We use Absolute Relative Error (Abs Rel) and $\delta < 1.25$ metrics to discard poor predictions caused by planar structures (c) or motion blur (d). (3) Alignment Quality Metrics: To address (e), we filter out images with insufficient projected LiDAR points or low spatial variance among those points, and constrain the output scale and shift parameters from the alignment algorithm.

\textbf{Large-Scale Dataset.} We apply this low-quality pseudo-label filtering pipeline to generate dense and accurate 3D point maps for a large-scale, mixed-domain dataset by aggregating five public autonomous driving benchmarks: Waymo~\cite{waymo}, nuScenes~\cite{nuscenes}, OpenScene~\cite{openscene}, DDAD~\cite{ddad}, and KITTI~\cite{kitti}. As illustrated in Figure~\ref{fig:gt}, our annotations capture complex scene geometry with high fidelity. By training and validating our model on this mixed dataset, we demonstrate its strong generalization ability and effectiveness across diverse, real-world driving scenarios.

\subsection{Training of \model}
\label{sec: method_train}
\paragraph{Training Losses.}
We train the \model in an end-to-end manner using a multi-task loss:
\vspace{-2mm}
\begin{equation}\label{eq:training_loss}
\mathcal{L}
=
\lambda \mathcal{L}_\text{epose}
+ \mathcal{L}_\text{pmap},
\vspace{-2mm}
\end{equation}
where both the ego pose loss $\mathcal{L}_\text{epose}$ and the point map loss $\mathcal{L}_\text{pmap}$ are metric-scaled. 
However, the numerical range of the point map values is significantly larger than that of the ego pose. 
We therefore set the ego pose loss weight $\lambda = 5.0$ to strike a balance. 
We describe each loss term in turn.

\begin{table*}[!t] \small
    \caption{\textbf{Quantitative 3D reconstruction results across diverse datasets.} * indicates that the predicted 3D point map is aligned with the sparse LiDAR depth using the Umeyama~\cite{umeyama2002least} algorithm to recover the metric scale. The inference time test is conducted on 128 images (16 frames $\times$ 8 views). Note that for CUT3R~\cite{cut3r} and StreamVGGT~\cite{streamvggt}, we report the cumulative inference time for streaming all images.}
\setlength{\tabcolsep}{8pt}
    \vspace{-3mm}
    \label{tab:mainpointmap}
    \centering
    \renewcommand{\tabcolsep}{5pt}     
    \begin{tabular}{ccc|cc|cc|cc|cc|c}
        \toprule
         & \multicolumn{2}{c}{\textbf{KITTI}} & \multicolumn{2}{c}{\textbf{NuScenes}} & \multicolumn{2}{c}{\textbf{Waymo}} & \multicolumn{2}{c}{\textbf{OpenScene}}  & \multicolumn{2}{c}{\textbf{DDAD}} & \textbf{}{Time} \\ 
         
        \cmidrule(lr){2-3} \cmidrule(lr){4-5} \cmidrule(lr){6-7} \cmidrule(lr){8-9} \cmidrule(lr){10-11} \cmidrule(lr){12-12}
        
         {\textbf{Method}} & {Acc $\downarrow$} & {Comp $\downarrow$} & {Acc $\downarrow$} & {Comp $\downarrow$} & {Acc $\downarrow$} & {Comp $\downarrow$} & {Acc $\downarrow$} & {Comp $\downarrow$} & {Acc $\downarrow$} & {Comp $\downarrow$} \\ 
        \midrule
        
        CUT3R*~\cite{cut3r} & 0.965 & 2.050 & 2.054 & 2.603 & 3.391 & 4.216 & 1.864 & 2.258 & 2.774 & 4.677 & $\sim$5.6s \\
        
        VGGT*~\cite{vggt} & 1.154 & 1.294 & 1.300 & 1.498 & 1.641 & 2.053 & 1.422 & 1.496 & 1.741 & 2.473 & $\sim$13.7s \\
        
        MapAnything~\cite{mapanything} & 1.880 & \textbf{1.014} & 4.499 & 4.886 & 10.205 & 8.494 & 3.353 & 4.303 & 8.015 & 8.493 & $\sim$5.8s \\ 

        StreamVGGT*~\cite{streamvggt} & 3.421 & 2.196 & 2.588 & 2.414 & 3.630 & 3.275 & 2.304 & 2.098 & 2.717 & 2.788 & $\sim$31.0s \\

        Driv3R*~\cite{fei2024driv3r} & 0.864 & 1.083 & 0.742 & 1.345 & \textbf{0.800} & \textbf{1.311} & 0.884 & 1.693 & 0.950 & 1.259 & $\sim$9.0s \\
                
        \model & \textbf{0.846} & 1.468 & \textbf{0.457} & \textbf{0.494} & 1.714 & 2.216 &\textbf{ 0.402} & \textbf{0.481} &\textbf{ 0.751} & \textbf{1.009} & \textbf{{$\sim$4.0s}}\\ 
        
        \bottomrule
    \end{tabular}
    \vspace{-3mm}
\end{table*}

\begin{table*}[t] \small
    \caption{\textbf{Quantitative ray depth results across diverse datasets.}}
    \vspace{-3mm}
    \label{tab:maindepthmap}
    \centering
    \renewcommand{\tabcolsep}{3pt}     
    \begin{tabular}{ccc|cc|cc|cc|cc}
        \toprule
         & \multicolumn{2}{c}{\textbf{KITTI}} & \multicolumn{2}{c}{\textbf{NuScenes}} & \multicolumn{2}{c}{\textbf{Waymo}} & \multicolumn{2}{c}{\textbf{OpenScene}}  & \multicolumn{2}{c}{\textbf{DDAD}} \\ 
         
        \cmidrule(lr){2-3} \cmidrule(lr){4-5} \cmidrule(lr){6-7} \cmidrule(lr){8-9} \cmidrule(lr){10-11}
        
         {\textbf{Method}} & {Abs Rel $\downarrow$} & { $\delta$\textless{}$1.25\uparrow$} & {Abs Rel $\downarrow$} & { $\delta$\textless{}$1.25\uparrow$} & {Abs Rel $\downarrow$} & { $\delta$\textless{}$1.25\uparrow$} & {Abs Rel $\downarrow$} & { $\delta$\textless{}$1.25\uparrow$} & {Abs Rel $\downarrow$} & { $\delta$\textless{}$1.25\uparrow$} \\ 
        \midrule
        
        CUT3R~\cite{cut3r} & 0.217 & 0.659 & 0.332 & 0.547 & 0.291 & 0.562 &  0.278 & 0.593 & 0.870 & 0.315 \\
        
        VGGT~\cite{vggt} & 0.158 & 0.801 & 0.243 & 0.729 & 0.176 & 0.811 & 0.241 & 0.719 & 0.613 & 0.476 \\
        
        MapAnything~\cite{mapanything} & 0.188 & 0.725 & 0.568 & 0.269 & 0.507 & 0.211 & 0.486 & 0.240 & 1.971 & 0.195 \\

        StreamVGGT~\cite{streamvggt} & 0.362 & 0.469 & 0.412 & 0.540 & 0.339 & 0.584 & 0.319 & 0.607 & 0.838 & 0.415 \\

        Driv3R~\cite{fei2024driv3r} & 0.164 & 0.784 & 0.189 & 0.721 & 0.168 & 0.770 & 0.188 & 0.740 & 0.185 & 0.740 \\

        \model & \textbf{0.136} & \textbf{0.849} & \textbf{0.069} & \textbf{0.953} & \textbf{0.106} &\textbf{ 0.921} & \textbf{0.049} & \textbf{ 0.971} & \textbf{0.152} & \textbf{0.837} \\

        \bottomrule
    \end{tabular}
    \vspace{-5mm}
\end{table*}

\begin{table}[t] \small
    \caption{\textbf{Quantitative ego pose results across diverse datasets.}}
    \vspace{-3mm}
    \label{tab:pose}%
\setlength{\tabcolsep}{2pt}
    \centering
    \resizebox{0.5\textwidth}{!}
    {
    \begin{tabular}{c|c|c|c|c|c}
    \toprule

    & KITTI & NuScenes & Waymo & OpenScene & DDAD \\
    {\textbf{Method}} & AUC@30 $\uparrow$ & AUC@30 $\uparrow$ & AUC@30 $\uparrow$ & AUC@30 $\uparrow$ & AUC@30 $\uparrow$ \\ 
    \midrule

    CUT3R~\cite{cut3r} & 51.8 & 43.5 & 50.1 & 34.7 & 48.6  \\
    
    VGGT~\cite{vggt} & \textbf{96.9} & \textbf{87.8} & \textbf{87.7} & 66.3 & 92.8  \\
    
    MapAnything~\cite{mapanything} & 90.6 & 85.0 & 82.8 & 65.6 & 87.0  \\
    
    StreamVGGT~\cite{streamvggt} & 95.8 & 86.2 & 85.6 & 74.1 & 91.9 \\
    
    \model & 87.6 & 86.5 & 86.4 & \textbf{74.7} & \textbf{95.1} \\
    \bottomrule
    \end{tabular}
    }
    \vspace{-7mm}
\end{table}

The ego pose loss $\mathcal{L}_\text{epose}$ supervises the predicted ego motion sequence $\mathcal{T}_{\text{ego}}$. 
We apply a standard L1 loss to the 7-dimensional pose representation, consisting of a 3D translation vector and a 4D rotation quaternion.
We compute the difference between the predicted pose $\hat{\mathbf{T}}_t$ and the ground truth $\mathbf{T}_t$ for each frame $t$. It can be described as below: 
\vspace{-2mm}
\begin{equation}
\label{eq:epose_loss}
\mathcal{L}_\text{epose} = \frac{1}{T} \sum_{t=1}^{T} \| \hat{\mathbf{T}}_t - \mathbf{T}_t \|_1.
\vspace{-2mm}
\end{equation}

We adopt the same loss functions as VGGT~\cite{vggt} for 3D point map supervision, which can be expressed as:
\vspace{-2mm}
\begin{align}
\mathcal{L}_\text{pmap} = &\sum_{t,n} \Big( \| \mathbf{\Sigma}_{t,n}^P \odot (\hat{\mathbf{P}}_{t,n} - {\mathbf{P}}_{t,n}) \|_2 \nonumber \\
& + \| (\nabla \hat{\mathbf{P}}_{t,n} - \nabla {\mathbf{P}}_{t,n}) \|_2 - \alpha \log \mathbf{\Sigma}_{t,n}^P \Big),
\vspace{-2mm}
\label{eq:pmap_loss}
\end{align}
where the sum $\sum_{t,n}$ iterates over all input images across $T$ frames and $N$ views. $\hat{\mathbf{P}}_{t,n}$ is the predicted 3D point map and ${\mathbf{P}}_{t,n}$ is the corresponding ground-truth. $\mathbf{\Sigma}_{t,n}^P$ is the additionally predicted per-pixel uncertainty map, measuring the uncertainty of the model's prediction for each pixel. $\odot$ refers to the channel-broadcast element-wise product. $\nabla$ denotes the 2D spatial gradient operator, and $\| \cdot \|_2$ is the L2 norm. The final term $- \alpha \log \mathbf{\Sigma}_{t,n}^P$ is a regularizer that encourages the model to be confident (with low uncertainty prediction), and we set the hyperparameter $\alpha = 2.0$.

\section{Experiments}

\label{sec: experiments}
We conduct extensive experiments to evaluate the performance of our \model. 
We provide dataset, implementation, and evaluation details in the appendix.

\subsection{3D Reconstruction and Depth Estimation}
As shown in \Cref{tab:mainpointmap} and \Cref{tab:maindepthmap}, we evaluate the 3D point reconstruction accuracy and ray depth accuracy (defined as the distance from a 3D point to the ego-vehicle center). Our analysis yields several key insights. First, \model demonstrates significantly superior 3D point and depth accuracy across driving datasets when compared to general visual geometry models~\cite{vggt, mapanything} and driving geometry models~\cite{fei2024driv3r}. This highlights the effectiveness of our architecture in reformulating the autonomous driving perception problem.

\begin{figure*}[!t]
\centering
\includegraphics[width=1\textwidth]{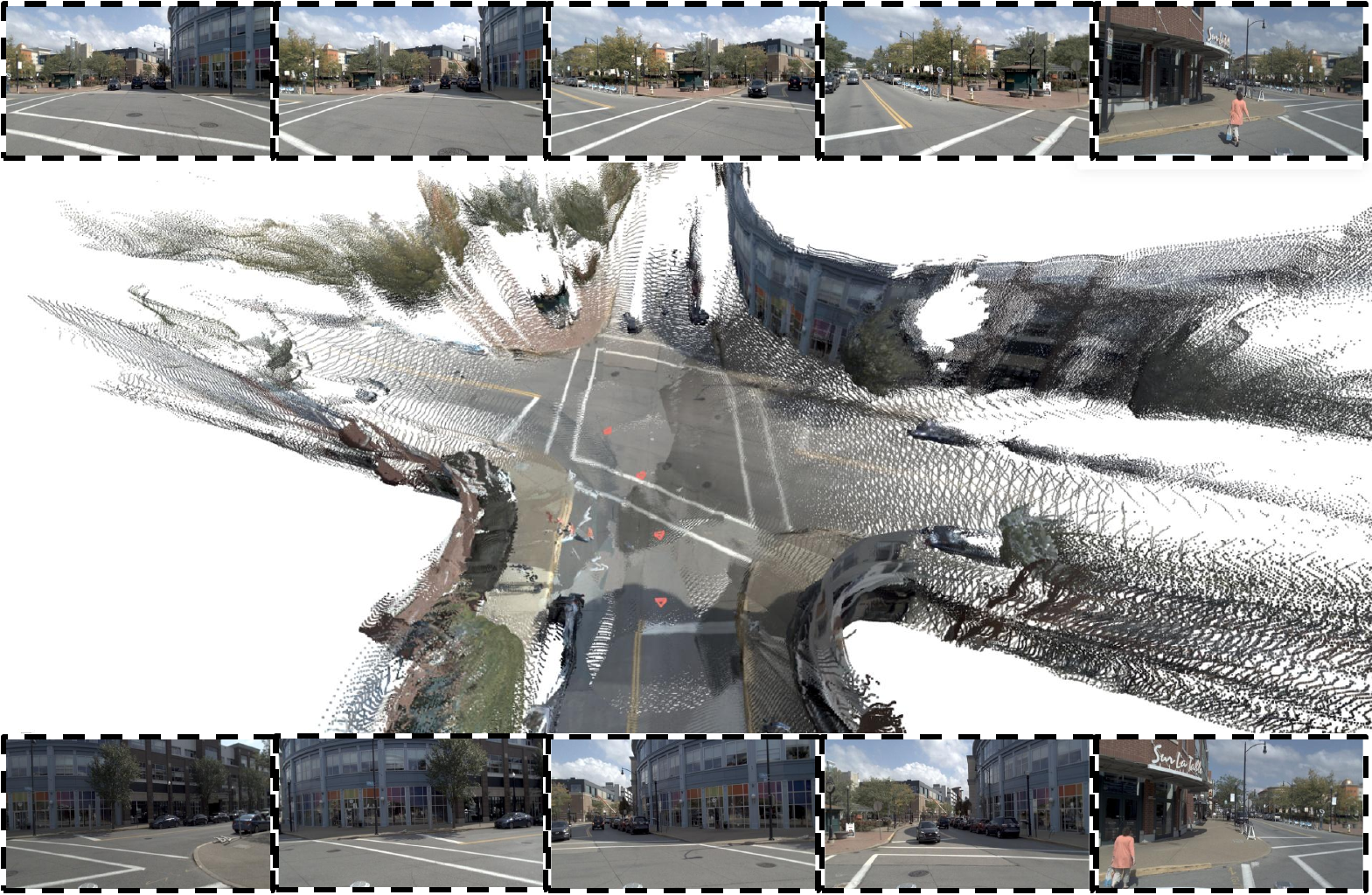}
\vspace{-7mm}
\caption{\textbf{Visualization of our DVGT on 3D point map reconstruction.}
}
\label{fig:qualitative_openscene}
\vspace{-6mm}
\end{figure*}

A crucial distinction is the handling of scale. Models like CUT3R~\cite{cut3r} and VGGT~\cite{vggt} necessitate post-hoc alignment with ground-truth point clouds to recover the metric scale. In contrast, \model directly predicts accurate, metric-scaled scene structure in a single, end-to-end forward pass. While MapAnything~\cite{mapanything} also predicts metric scale without post-processing, its reconstruction error is substantially larger (e.g., 4.303 vs. 0.481 for \model on OpenScene~\cite{openscene}), underscoring the superior accuracy of our approach.

Furthermore, we note that \model's performance on Waymo~\cite{waymo} is less competitive. We attribute this to disproportionate data sampling: Waymo~\cite{waymo} was assigned the same weight as smaller datasets despite being $5\times$ larger during training. Unlike nuScenes~\cite{nuscenes}, which benefits from a distribution similar to the dominant OpenScene~\cite{openscene} data, Waymo lacks this alignment. We expect that optimizing the sampling weights of different datasets will bridge this gap.

\subsection{Ego-Pose Estimation}
In addition to scene geometry, \model jointly estimates ego pose. Table~\ref{tab:pose} presents the results for pose prediction. Our model achieves better results on OpenScene~\cite{openscene} and DDAD~\cite{ddad}, while attaining comparable performance to VGGT~\cite{vggt} on nuScenes~\cite{nuscenes} and Waymo~\cite{waymo}. This demonstrates that \model effectively serves as a comprehensive visual geometry model, capable of accurately estimating joint scene structure and ego-vehicle pose.

We note that \model's pose prediction performance on KITTI~\cite{kitti} is slightly lower. We attribute this to KITTI's high-overlap dual-camera setup, which limits full 3D and ego-motion understanding compared to surround-view data.

\begin{table}[t]
    \caption{\textbf{Comparison with driving models for depth estimation compared with LiDAR GT on the nuscenes~\cite{nuscenes} dataset.}}
    \vspace{-3mm}
    \label{tab:comparison_lidar}
    \centering
    \resizebox{0.495\textwidth}{!}{
    \begin{tabular}{c|c|cc}
        \toprule
        
         Method & Scale Method & Abs Rel $\downarrow$ & $\delta<1.25\uparrow$ \\ 
        \midrule
        MonoDepth2~\cite{monodepth2} & Median Scaling & 0.29 & 0.64 \\
        
        SurroundDepth~\cite{surrounddepth} & SfM Pretrain & 0.28 & 0.66 \\

        R3D3~\cite{r3d3} & Extrinsic & 0.25 & 0.73 \\

        SelfOcc~\cite{selfocc} & Pose GT & 0.23 & 0.75 \\

        Dist4D~\cite{dist4d} & Pose GT & 0.39 & 0.58 \\

        OmniNWM~\cite{omninwm} & Pose GT & 0.23 & 0.81 \\

        \textbf{DVGT} & None & \textbf{0.13} & \textbf{0.86} \\
        
        \bottomrule
    \end{tabular}
    }
    \vspace{-7mm}
\end{table}

\subsection{Comparison with Driving Models}

We perform a comparison with existing driving models for depth estimation on the widely used nuScenes~\cite{nuscenes} dataset. Our model directly reconstructs 3D point maps, whereas most driving models predict depth maps that are evaluated against sparse LiDAR ground truth. For a fair comparison, we convert our 3D point map prediction into depth maps and compute the same metrics.

Table~\ref{tab:comparison_lidar} shows that our \model achieves state-of-the-art performance, obtaining the best scores on both Abs Rel and $\delta<1.25$ accuracy. Note that while other methods require post-processing like median scaling or depend on ground-truth camera poses for scale recovery, \model operates without any post-alignment or pose information. This demonstrates the effectiveness of our model in recovering metrically accurate 3D geometry directly from images.

\subsection{Experimental Analysis}
\textbf{Ablation Study on Scene Scale.}
A significant challenge in autonomous driving perception is the vast dynamic range of scene geometry, with distances often exceeding 100 meters. Directly regressing large-valued 3D coordinates can introduce numerical instability during training, as model parameters may be pushed to large magnitudes, potentially leading to performance degradation.

\begin{table}[t] \small
    \caption{\textbf{Ablation study on ground truth coordinate normalization on the nuScenes~\cite{nuscenes} dataset.} 1, 10, and 100 denote linearly scale dividing by 1, 10, and 100, respectively. And asinh denotes the arcsinh function for non-linearly scaling.}
    \label{tab:ab_scale}
    \vspace{-3mm}
    \centering
    \renewcommand{\tabcolsep}{3.pt}
    \begin{tabular}{cccccc}
        \toprule
        
         {\textbf{Method}} & {Acc $\downarrow$} & {Comp $\downarrow$} & {Abs Rel $\downarrow$} & { $\delta$\textless{}$1.25\uparrow$} & AUC@30 $\uparrow$ \\ 
        \midrule
        
        1 (base) & 1.584 & 1.424 & 0.261 & 0.676 & 68.4 \\
        
        10  & \textbf{1.349} & \textbf{1.053} & \textbf{0.195} & \textbf{0.756} & 79.8 \\
        
        100  & 1.646 & 1.431 & 0.257 & 0.694 & 80.7 \\
                
        asinh  & 1.411 & 1.390 & 0.222 & 0.719 & \textbf{80.8} \\
        
        \bottomrule
    \end{tabular}
    \vspace{-7mm}
\end{table}

Therefore, we investigate strategies for scaling the target 3D points to a more stable numerical range for regression. As shown in \Cref{tab:ab_scale}, we conduct an ablation study comparing three distinct scaling methods:
\begin{itemize}
    \item Linear Scaling (10x): Linearly dividing all target coordinates by 10.
    \item Linear Scaling (100x): Linearly dividing all target coordinates by 100.
    \item Non-linear Scaling (arcsinh): Applying the arcsinh function to non-linearly compress the coordinate range.
\end{itemize}

\begin{table}[!t] \small
    \caption{\textbf{Ablation study on attention mechanisms on the nuScenes~\cite{nuscenes} dataset.}
    We compare combinations of different attention components: 
    \textbf{G} (Global Attention), 
    \textbf{L} (Intra-View Local Attention), 
    \textbf{S} (Cross-View Spatial Attention ), 
    \textbf{T} (Cross-Frame Temporal Attention), and 
    \textbf{TE} (Temporal Positional Embedding).
    }    \label{tab:ab_attn}
    \vspace{-3mm}
    \centering
    \renewcommand{\tabcolsep}{2pt}
    \resizebox{0.475\textwidth}{!}{
    \begin{tabular}{ccccccc}
        \toprule
        
         {\textbf{Method}} & {Acc $\downarrow$} & {Comp $\downarrow$} & {Abs Rel $\downarrow$} & { $\delta$\textless{}$1.25\uparrow$} & AUC@30 $\uparrow$  & Time \\ 
        \midrule
        
        L+G & \textbf{1.131} & \textbf{1.129} & \textbf{0.178} & \textbf{0.789} & 74.6 & $\sim$8.2s \\
        
        L+S+T & 1.584 & 1.424 & 0.261 & 0.676 & 68.4 & \textbf{$\sim$4.0s} \\
        
        L+S+T+TE & 1.458 & 1.139 & 0.227 & 0.725 & \textbf{77.6} & \textbf{$\sim$4.0s} \\
        
        \bottomrule
    \end{tabular}
    }
    \vspace{-7mm}
\end{table}

\begin{figure*}[t]
\centering
\includegraphics[width=1.0\textwidth]{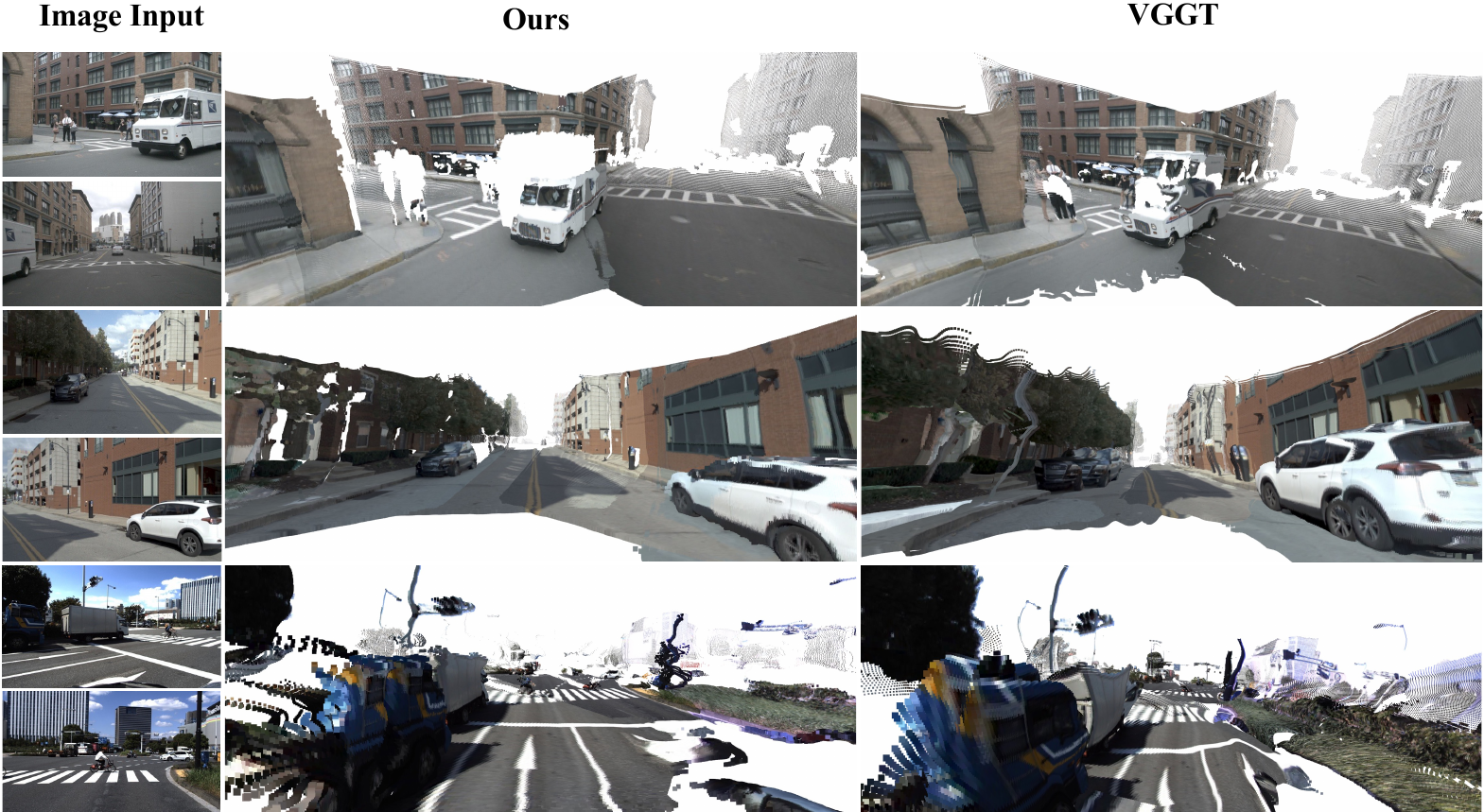}
\vspace{-7mm}
\caption{\textbf{Comparison of our DVGT with VGGT on 3D metric-scaled point map reconstruction.} Notably, VGGT performed post-alignment with LiDAR points GT while our DVGT eliminates the need for any post-processing.
}
\label{fig:comparison}
\vspace{-5mm}
\end{figure*}

The experimental results indicate that linear scaling by 10x achieves the best performance. We analyze that scaling by 100x makes the scene scale too small, which degrades precision for near-field geometry. The $\text{arcsinh}$ transformation offers an adaptive scaling (less compression for near), but this non-linear operation may distort the inherent geometric structure, resulting in sub-optimal performance.

\textbf{Ablation Study on Attention Design.}
We perform an ablation study on the attention mechanism within our geometry transformer in Table~\ref{tab:ab_attn}. We compare our decomposed attention mechanism against a standard global attention baseline, where all tokens interact with all others. Our method yields a significant improvement in efficiency and a faster inference speed. This decomposition, however, introduces a minor performance trade-off. To mitigate this, we incorporate temporal positional embeddings, providing the model with explicit sequential information. This addition successfully narrows the performance gap, allowing our model to achieve a compelling balance between high performance and computational efficiency.

\textbf{Qualitative Analysis.}
We provide qualitative visualizations to further analyze the capabilities of our model. In \Cref{fig:qualitative_openscene}, the visualization confirms that the ego-pose is accurately estimated. Furthermore, the model exhibits high multi-frame and multi-view consistency for static elements, like road surfaces and buildings. Dynamic objects, such as pedestrians and vehicles at the intersection, are also faithfully reconstructed, while fine-grained details, including roadside trees, are well-preserved.

\Cref{fig:comparison} shows that VGGT struggles with both static elements (e.g., trees, lanes) and dynamic objects (e.g., vehicles, pedestrians), often failing to capture accurate geometry. In contrast, our \model, leveraging dense annotations and large-scale training, successfully recovers scene geometry while maintaining high efficiency.

\section{Conclusion}
\label{sec: conclusion}
In this paper, we introduced \model, a universal visual geometry model for autonomous driving. 
\model generates a dense, metric-scaled 3D global point map from unposed images, enabling an accurate and complete understanding of scene geometry. Its 3D prior-free transformer architecture enables adaptability to different camera configurations, overcoming a major bottleneck for scalability. To validate our approach and facilitate future research, we also constructed a large-scale driving dataset with dense 3D point annotations. Experiments demonstrate that \model significantly outperforms existing methods, paving the way for more robust and versatile vision-centric driving systems.

\section*{Acknowledgements}
This work was supported in part by the National Natural Science Foundation of China under Grant 62576188, Grant 62336004, Grant 62321005, and Grant 62125603, in part by the Beijing Natural Science Foundation under Grant L247009, and in part by the Beijing National Research Center for Information Science and Technology.

{
    \small
    \bibliographystyle{ieeenat_fullname}
    \bibliography{main}

\appendix
\clearpage

\twocolumn[{%
\renewcommand\twocolumn[1][]{#1}%
\begin{center}
    \centering
    \includegraphics[width=0.95\linewidth]{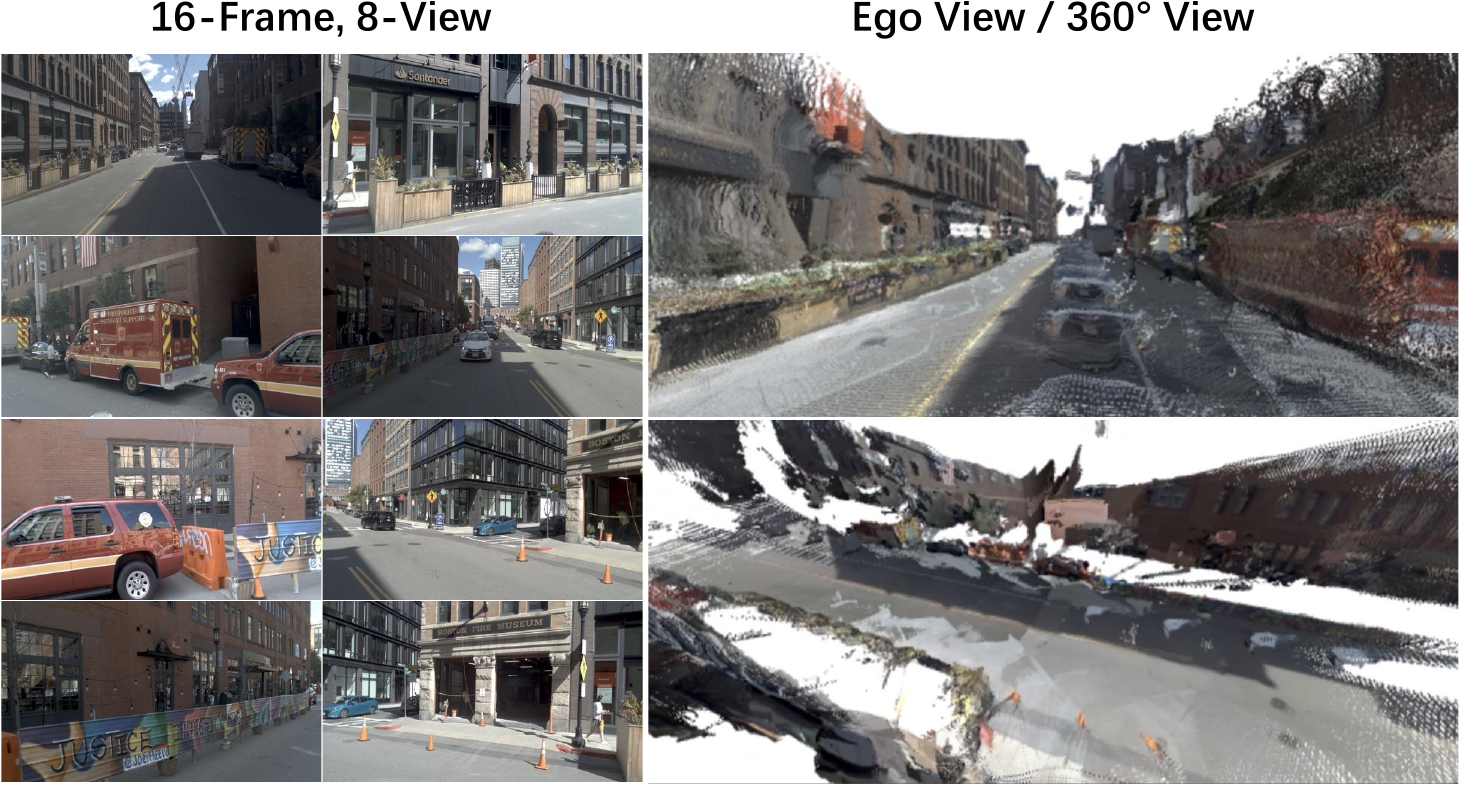}
    \vspace{-2mm}
    \captionof{figure}{\textbf{Video demonstration of our \model's reconstruction of scene geometry from images on the validation set.}  
}
\label{fig:demo}
\end{center}%
}]

\begin{table*}[!ht]
    \vspace{-3mm}
    \label{tab:mainpointmap}
    \centering
    \caption{Detailed statistics of the datasets used in our experiments. All temporal statistics are reported at a 2Hz sampling rate.}
    \vspace{-3mm}
    \footnotesize 
    \setlength{\tabcolsep}{9pt}
    \label{tab:dataset_info}
    \begin{tabular}{c|c|c|c|c|c|c|c}
        \toprule
        \textbf{Dataset} & \textbf{Train Scenes} & \textbf{Test Scenes} & \textbf{Min Frames} & \textbf{Max Frames} & \textbf{Avg Frames} & \textbf{Avg Aspect Ratio} & \textbf{Num of Views} \\ 
        \midrule
        nuScenes  & 700   & 150  & 32 & 41   & 40 & 1.77 & 6 \\
        KITTI     & 138   & 13   & 2  & 1033 & 62 & 3.31 & 2 \\
        OpenScene & 19376 & 2026 & 1  & 41   & 38 & 1.77 & 8 \\
        Waymo     & 798   & 202  & 34 & 40   & 40 & 1.77 & 5 \\
        DDAD      & 150   & 50   & 10 & 20   & 17 & 1.59 & 6 \\ 
        \bottomrule
    \end{tabular}
    \vspace{-6mm}
\end{table*}

\section*{\Large Appendix}



\section{Implementation Details}

\textbf{Architecture}.
We utilize a ViT-L model pretrained by DINOv3~\cite{dinov3} as the image encoder. The subsequent geometry transformer is composed of $L=24$ attention blocks, bringing the model size to approximately 1.7 billion parameters. Each block consists of three specialized layers: an intra-view local attention layer, a cross-view spatial attention layer, and a cross-frame temporal attention layer. Following the ViT-L configuration~\cite{dinov3}, each attention layer is set to a feature dimension of 1024 with 16 heads. To enhance training stability, we incorporate QKNorm~\cite{qknorm} and LayerScale~\cite{layerscale} (initialized at 0.01) into each layer. For dense prediction, we follow the design in~\cite{depthanything2} and feed tokens from the 4th, 11th, 17th, and 23rd blocks into a DPT~\cite{dpt} decoder for upsampling.

\textbf{Training}. 
We train our model on a mixture of public datasets using the AdamW~\cite{adamw} optimizer for $160$K iterations. We employ a cosine learning rate scheduler with a peak learning rate of $10^{-4}$ and a linear warmup of $8$K iterations. To ensure training stability and efficiency, we apply gradient norm clipping with a threshold of $1.0$ and leverage bfloat16 precision alongside gradient checkpointing. The training process takes about six days on 64 H20 GPUs.

During training, we construct each batch (batch size of 1) by first sampling a dataset based on its weight, followed by a random scene. From this scene, we dynamically sample views (from 2 to the maximum available) and frames to yield a total of 48 images per batch. Input images are isotropically resized to a long-edge resolution of 518 pixels. We then apply a central crop on the shorter edge to a random size between 144 and 224 pixels (ensuring divisibility by the 16-pixel patch size), resulting in aspect ratios between 1.5 and 3.3. Following VGGT~\cite{vggt}, we apply aggressive per-frame augmentations—including color jittering, Gaussian blur, and grayscale conversion—to improve robustness against varying lighting conditions.

\section{Evaluation Metrics}
We evaluate our method's performance on two primary tasks: 3D point map reconstruction and ego-pose estimation.
For 3D point map reconstruction, we adopt two sets of metrics.
First, following prior works~\cite{vggt, pi3, mapanything}, we measure overall geometric quality using \textbf{Accuracy} and \textbf{Completeness}. 
Formally, let $\mathcal{P} = \{\mathbf{p}_i\}$ and $\mathcal{G} = \{\mathbf{g}_j\}$ denote the sets of valid points from the predicted point map $\mathbf{P}_{pred}$ and the ground truth $\mathbf{P}_{gt}$, respectively. These metrics are calculated as:
\begin{equation}
    \begin{gathered}
        \textbf{Accuracy} = \frac{1}{|\mathcal{P}|} \sum_{\mathbf{p} \in \mathcal{P}} \min_{\mathbf{g} \in \mathcal{G}} \|\mathbf{p} - \mathbf{g}\|_2, \\
        \textbf{Completeness} = \frac{1}{|\mathcal{G}|} \sum_{\mathbf{g} \in \mathcal{G}} \min_{\mathbf{p} \in \mathcal{P}} \|\mathbf{g} - \mathbf{p}\|_2,
    \end{gathered}
\end{equation}
where $\|\cdot\|_2$ denotes the $L_2$ distance, quantifying the proximity between the prediction and the underlying scene geometry.

Second, we evaluate the ray depth accuracy using the \textbf{Absolute Relative error (Abs Rel)} and \textbf{threshold accuracy ($\delta < 1.25$)}.
Here, ray depth refers to the distance from a 3D point to the ego-vehicle center of its corresponding frame, which measures the local structure of the 3D point map reconstruction.
Let $\Omega$ be the set of valid pixels with available ground truth, and $u$ index a pixel within $\Omega$. Given the predicted ray depth $\mathbf{D}_{pred}$ and ground truth $\mathbf{D}_{gt}$, the metrics are defined as:
\begin{equation}
    \begin{gathered}
        \text{Abs Rel} = \frac{1}{|\Omega|} \sum_{u \in \Omega} \frac{| \mathbf{D}_{pred}(u) - \mathbf{D}_{gt}(u) |}{\mathbf{D}_{gt}(u)}, \\
        \text{Acc}_{\delta} = \frac{1}{|\Omega|} \sum_{u \in \Omega} \mathbb{I}\left( \max\left( \frac{\mathbf{D}_{pred}(u)}{\mathbf{D}_{gt}(u)}, \frac{\mathbf{D}_{gt}(u)}{\mathbf{D}_{pred}(u)} \right) < \delta \right),
    \end{gathered}
\end{equation}
where $\mathbb{I}(\cdot)$ is the indicator function that equals 1 if the condition holds and 0 otherwise.

For ego-pose estimation, following \cite{wang2023posediffusion}, we report the \textbf{Area Under the Curve (AUC)} of the pose accuracy at a $30^\circ$ threshold. 
Specifically, we first compute the Relative Rotation Accuracy (RRA) and Relative Translation Accuracy (RTA) for all frame pairs. RRA measures the geodesic distance between the predicted and ground truth rotation matrices, while RTA evaluates the angular deviation between the translation vectors.
The accuracy at a specific threshold $\tau$ is defined as the percentage of pairs satisfying $\max(\text{RRA}, \text{RTA}) < \tau$. 
The final AUC@30 is obtained by integrating this accuracy over the threshold range:
\begin{equation}
    \text{AUC}@30^\circ = \frac{1}{30} \int_{0}^{30} \text{Acc}(\tau) \, d\tau,
\end{equation}
where $\text{Acc}(\tau)$ represents the fraction of camera pairs with both angular errors smaller than $\tau$.

\section{Video Demonstration}
Figure~\ref{fig:demo} shows a sampled image from the video demo that demonstrates our model's reconstruction of 3D scene geometry on the validation set. Given multi-frame, multi-view images as input, \model generates dense point maps which recover scene geometry with high fidelity and consistency, validating the effectiveness of our method.

\section{Dataset Details}
\label{sec:dataset_details}
We utilize five diverse datasets for training and evaluation: Waymo~\cite{waymo}, nuScenes~\cite{nuscenes}, OpenScene~\cite{openscene}, DDAD~\cite{ddad}, and KITTI~\cite{kitti}. Detailed statistics for each dataset are summarized in Table~\ref{tab:dataset_info}. All data sequences are downsampled to a temporal frequency of 2Hz, and the frame counts reported in the table are calculated based on this sampling rate.

During the training phase, the sampling ratio across datasets is set as \textit{nuScenes : KITTI : OpenScene : Waymo : DDAD} = $6 : 5 : 77 : 6 : 6$. For each training iteration, we first select a dataset according to these weights and sample a batch from it. To ensure robust feature learning across various sensor configurations, we implement a dynamic sampling strategy:
\begin{itemize}
    \item A target aspect ratio is randomly sampled from the range $[1.5, 3.3]$.
    \item The number of camera views is randomly selected from $[2, 8]$.
    \item Given a hardware constraint of 48 images per GPU per iteration, we determine the maximum possible sequence length $T_{max}$.
    \item A specific frame number is then sampled from $[2, T_{max}]$, and the final batch size is calculated to saturate the GPU memory efficiency.
\end{itemize}

}


\end{document}